\documentclass[sigconf]{acmart}
\AtBeginDocument{%
  }

\setcopyright{acmlicensed}
\copyrightyear{2018}
\acmYear{2018}
\acmDOI{XXXXXXX.XXXXXXX}
\acmConference[Conference acronym 'XX]{Make sure to enter the correct
  conference title from your rights confirmation email}{June 03--05,
  2018}{Woodstock, NY}
\acmISBN{978-1-4503-XXXX-X/2018/06}

\usepackage{enumitem}
\usepackage{booktabs}
\usepackage{multirow}
\usepackage{listings}

\usepackage{amsmath}
\usepackage{bm}
\usepackage{mathtools}
\usepackage{amsthm}
\usepackage{pifont}
\usepackage{multirow}
\usepackage{subcaption}

\newtheorem{theorem}{Theorem}
\newtheorem*{theorem*}{Theorem}

\newtheorem{definition}{Definition}
\newtheorem{assumption}{Assumption}

\DeclareMathOperator*{\argmax}{arg\,max}
\DeclareMathOperator*{\argmin}{arg\,min}

\newcommand{\hlr}[1]{\colorbox{red!10}{#1}}          
\newcommand{\hla}[1]{\colorbox{green!10}{#1}}        
\newcommand{\hls}[1]{\colorbox{blue!8}{#1}}          

\begin{document}

\title{Structure-R1: Dynamically Leveraging Structural Knowledge in LLM Reasoning through Reinforcement Learning}

\author{Junlin Wu}
\affiliation{
  \institution{Washington University in St. Louis} 
  \city{St. Louis}
  \state{MO}
  \country{USA}}
\email{junlin.wu@wustl.edu}

\author{Xianrui Zhong}
\affiliation{%
  \institution{University of Illinois at Urbana-Champaign}
  \city{Urbana}
  \state{IL}
  \country{USA}}
  \email{xzhong23@illinois.edu}


\author{Jiashuo Sun}
\affiliation{%
  \institution{University of Illinois at Urbana-Champaign}
  \city{Urbana}
  \state{IL}
  \country{USA}}
\email{jiashuo5@illinois.edu}

\author{Bolian Li}
\affiliation{%
  \institution{Purdue University}
  \city{West Lafayette}
  \state{IN}
  \country{USA}}
\email{li4468@purdue.edu}

\author{Bowen Jin}
\affiliation{%
  \institution{University of Illinois at Urbana-Champaign}
  \city{Urbana}
  \state{IL}
  \country{USA}}
\email{bowenj4@illinois.edu}

\author{Jiawei Han}
\affiliation{%
  \institution{University of Illinois at Urbana-Champaign}
  \city{Urbana}
  \state{IL}
  \country{USA}}
\email{hanj@illinois.edu}

\author{Qingkai Zeng}
\authornote{Corresponding Author}
\affiliation{%
  \institution{University of Notre Dame}
  \city{Notre Dame}
  \state{IN}
  \country{USA}}
\email{qzengnkcs@gmail.com}

\settopmatter{printacmref=false} 
\renewcommand\footnotetextcopyrightpermission[1]{} 



\begin{abstract}

Large language models (LLMs) have demonstrated remarkable advances in reasoning capabilities.
However, their performance remains constrained by limited access to explicit and structured domain knowledge. 
Retrieval-Augmented Generation (RAG) addresses this by incorporating external information as context to augment reasoning.
Nevertheless, traditional RAG systems typically operate over unstructured and fragmented text, resulting in low information density and suboptimal reasoning. 
To overcome these limitations, we propose \textsc{Structure-R1}, a novel framework that transforms retrieved content into structured representations optimized for reasoning. 
Leveraging reinforcement learning, \textsc{Structure-R1} learns a content representation policy that dynamically generates and adapts structural formats based on the demands of multi-step reasoning. 
Unlike prior methods that rely on fixed schemas, our approach adopts a generative paradigm capable of producing task-specific structures tailored to individual queries. 
To ensure the quality and reliability of these representations, we introduce a self-reward structural verification mechanism that checks whether the generated structures are both correct and self-contained. 
Extensive experiments on seven knowledge-intensive benchmarks show that \textsc{Structure-R1} consistently achieves competitive performance with a 7B-scale backbone model and matches the performance of much larger models. 
Additionally, our theoretical analysis demonstrates how structured representations enhance reasoning by improving information density and contextual clarity. 
Our code and data are available at: \url{https://github.com/jlwu002/sr1}.
\end{abstract}




 \maketitle

\section{Introduction}
Recent advancements in large language models (LLMs) have demonstrated substantial improvements in natural language understanding~\cite{zhang2023pre,xu2024large} and generation~\cite{achiam2023gpt, team2024qwen2,guo2025deepseek}. These capabilities have led to their widespread deployment in a range of complex, real-world applications across diverse domains~\cite{peng2023study, wu2023bloomberggpt,zhang2024comprehensive}. 
Despite these impressive capabilities, LLMs still struggle with tasks that require complex reasoning or domain-specific expertise~\cite{huang2025survey}, such as those in scientific~\cite{zhang2023pre,zhang2024comprehensive}, medical~\cite{lucas2024reasoning}, or legal domains~\cite {zhou2024unlocking,yao2025intelligent}. 
A key limitation is their lack of explicit and up-to-date domain knowledge, which constrains their reasoning performance. To mitigate this issue, users are often required to manually provide external context or supporting materials. One promising approach is Retrieval-Augmented Generation (RAG)~\cite{lewis2020retrieval, chung2024scaling, asai2023self, jin2025search}, which enhances LLMs by integrating external knowledge retrieval into the inference process. This enables models to dynamically access and incorporate relevant information from large-scale sources, thereby improving their effectiveness on domain-specific tasks.

\begin{figure}[!t]
    \centering
\includegraphics[width=0.75\linewidth]{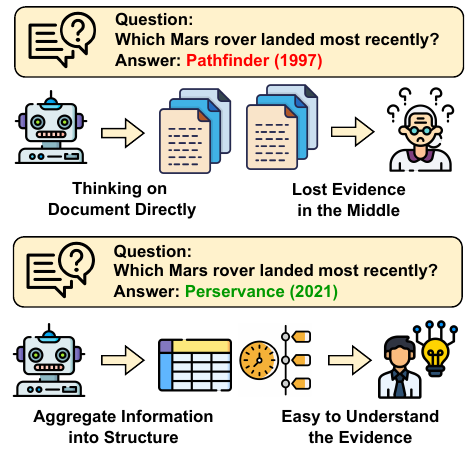}
    \vspace{-0.1in}
    \Description{}
    \caption{Scattered information across documents can hinder reasoning. Structuring content into formats like tables or timelines helps consolidate key details and improve accuracy.}
    \label{fig:motivation}
    \vspace{-0.2in}
\end{figure}

While RAG establishes a foundation for connecting LLMs with external knowledge, they typically process the retrieved information as segmented text chunks. 
However, due to limitations in the retriever’s ability to fully understand complex queries and the information capacity constraint of the queries themselves, the retrieved content tends to contain fragmented, loosely related pieces scattered across multiple documents~\cite{listructrag,wang2024leave, jiang2025retrieval}. 
As shown in Figure~\ref{fig:motivation}, the density of relevant content within any individual chunk is low, which hinders the effectiveness of the models on the downstream tasks. 
More specifically, LLMs do not perform complex reasoning by simply reading the retrieved raw text. 
Instead, they suffer from ``lost in the middle''~\cite{liu2024lost}, overlooking critical information due to the presence of excessive or redundant content introduced by the retriever in an attempt to maximize knowledge coverage.

Motivated by prior work~\cite{yu2022survey,jiang2025retrieval,shi2025hypercube} on leveraging structural knowledge to enhance knowledge understanding in LLMs, an emerging direction is to process retrieved content into structured forms. These forms can increase the \textit{density} and \textit{clarity} of relevant information by organizing key elements in a more coherent and task-aligned manner compared to free-form text. Existing graph-based RAG systems~\cite{edge2024local, guo2024lightrag, gutierrez2025rag, luo2025hypergraphrag} enhance retrieval efficiency and enable knowledge-driven generation by constructing knowledge graphs from retrieved documents. These graphs capture inter-entity relationships, allowing the model to reason over contextual dependencies rather than treating entities in isolation. StructRAG~\cite{listructrag} expands beyond homogeneous knowledge graphs to support heterogeneous structured formats such as tables, catalogs, and algorithms. In details,it selects the most suitable structure from a predefined set, builds the structure from retrieved content, and guides the LLM to perform reasoning over it through prompt-based generation. As illustrated in Figure~\ref{fig:motivation}, consider the query question \textit{``Which Mars rover landed most recently?''}  Instead of reasoning directly over raw retrieved text, structuring the relevant information into a table helps concentrate and organize key content, making it easier for the model to interpret and reason over.

While structured forms help organize retrieved documents for better comprehension, real-world reasoning is often complex and typically unfolds step-by-step through a logical chain~\cite{wei2022chain}. Consequently, the optimal structural format for representing knowledge may vary across different reasoning steps. Reconsider the example in Figure~\ref{fig:motivation}: after organizing relevant information into a table that maps missions to their respective landing dates, the system may further transform this table into a timeline format. This transformation facilitates easier identification of the most recent landing event. In addition, although LLMs have shown promise in information extraction~\cite{xu2024large, zeng2024chain, zeng2024codetaxo}, accurately identifying and extracting correct structural knowledge from retrieved documents remains a substantial challenge. To address these two challenges: (1) The dynamic need for adaptive structural formats during multi-step reasoning; and (2) the difficulty of reliable structured knowledge extraction from raw documents, we proposed e proposed \textsc{Structure-R1}, a novel framework for knowledge-intensive reasoning tasks.

\textsc{Structure-R1} leverages reinforcement learning to optimize a content representation policy that transforms retrieved information into structured knowledge representations. These structured forms facilitate more efficient access to relevant information, enabling LLMs to perform more precise and coherent reasoning over complex inputs. Furthermore, \textsc{Structure-R1} adopts a generative paradigm, freeing it from the constraints of predefined structural schemas. When existing schemas are insufficient, \textsc{Structure-R1} is capable of generating novel structural formats tailored to the reasoning requirements. To ensure the correctness and relevance of the generated structural knowledge representations, we propose a modified two-stage rollout strategy based on the traditional GRPO~\cite{shao2024deepseekmath} framework.  Specifically, we design a dual evaluation setup to verify whether the generated structures are both self-contained and sufficient for reasoning. For each query, the LLM is prompted to reason under two conditions: (1) using both the retrieved documents and the generated structure, and (2) using the generated structure alone, without access to the original documents. This setup allows us to assess whether the structured representation itself effectively encapsulates the necessary knowledge for answering the question. We evaluate the proposed \textsc{Structure-R1} on seven knowledge-intensive benchmarks, covering both in-domain and out-of-domain settings. 
Our method achieves state-of-the-art performance among models using a 7B-scale backbone and even matches or outperforms larger models such as GPT-4o-mini on multiple benchmarks. To further understand the effectiveness of \textsc{Structure-R1}, we also provide a theoretical analysis illustrating how structured formats enhance knowledge-intensive reasoning from an information density perspective.

In summary, this study makes the following contributions:
\begin{itemize}[noitemsep, topsep=0pt]
    \item We propose \textsc{Structure-R1}, a novel framework that leverages reinforcement learning to optimize a content representation policy, enabling the transformation of retrieved textual information into structured knowledge representations.
    \item To ensure the correctness and self-containment of the generated structures, we introduce a self-rewarding verification module that improves the quality and reliability of the structured outputs.
    \item Extensive experiments across seven benchmarks demonstrate that \textsc{Structure-R1} significantly improves performance on knowledge-intensive reasoning tasks in both in-domain and out-of-domain settings. Additionally, a theoretical analysis is provided to support the effectiveness of our approach.
\end{itemize}

\indent \textbf{Scope and Limitation.} This work focuses on learning a policy that enables the model to represent retrieved documents in the most suitable and accurate structured formats, thereby enhancing its reasoning capability. Unlike other related efforts~\cite{jin2025search, luo2025graph,yu2025graphrag} that also employ reinforcement learning or graph-based knowledge within RAG systems, our approach does not aim to improve retrieval quality itself. Instead, the key distinction lies in our objective: we assume the retrieved content is given and concentrate on its better utilization through transformation into structured format representations.

\section{Problem Definition}
In this section, we present the key concepts and formally define the knowledge-intensive reasoning tasks as follows:

\begin{definition}[Retrieved-context Reasoning Task]
    Given a query $q$, and a set of retrieved documents in plain-text $\mathcal{D}_q = \{d_1,d_2,\dots, d_n \}$  containing potentially relevant evidence, the Retrieved-context reasoning task aims to learn a model $f:(q, \mathcal{D}_q) \rightarrow a$, that  performs multi-step reasoning over $\mathcal{D}_q$ to produce a target output $a$.
\end{definition}

While this formulation offers a general framework for knowledge-intensive reasoning, it typically treats retrieved documents as unstructured text. However, recent studies~\cite{listructrag} show that flattening documents into plain text chunks can obscure structural cues and semantic relationships. This often leads LLMs to overlook or misinterpret key information, increasing the risk of hallucinated or inaccurate responses. To mitigate this, we propose enhancing retrieved-context reasoning by converting documents into structured representations that better preserve task-relevant content. We focus on question answering (QA) tasks, assuming a fixed set of retrieved documents per query. Under this setting, we define our task as follows:

\begin{definition}[Structure Enhanced QA Task]
    Given a question $q$, a set of retrieved documents in $\mathcal{D}_q = \{d_1,d_2,\dots, d_n \}$, and a predefined set of structural format type $\mathcal{S}=\{s_1, s_2, \dots, s_m\}$, the dynamic structure enhanced QA Task aims to learn a model $f:(q, \mathcal{D}_q, \mathcal{S}) \rightarrow (a, \mathcal{S'})$, that leverages corpus $\mathcal{D}_q$ and outputs the accurate answer $a$, along with a selected or newly constructed subset of structure formats $\mathcal{S}' \subseteq \mathcal{S} \cup \mathcal{S}^+$. Here, $\mathcal{S}^+$ denotes the set of novel structure formats dynamically generated by the model when the predefined set $\mathcal{S}$ is insufficient for the query.
\end{definition}

\section{Methodology\label{sec:method}}
In this section, we present \textsc{Structure-R1}, an R1-inspired framework that integrates structured prompting with reinforcement learning to enhance reasoning in LLMs. 
Our approach extends the R1 paradigm to explicitly separate reasoning, reformatting, and answering through format-aware control tokens. 

\subsection{Training Template for Structure-R1}
\label{subsec_train_temp}
\subsubsection{Task description}
\begin{figure*}[t!]
    \centering
\includegraphics[width=0.8\linewidth]{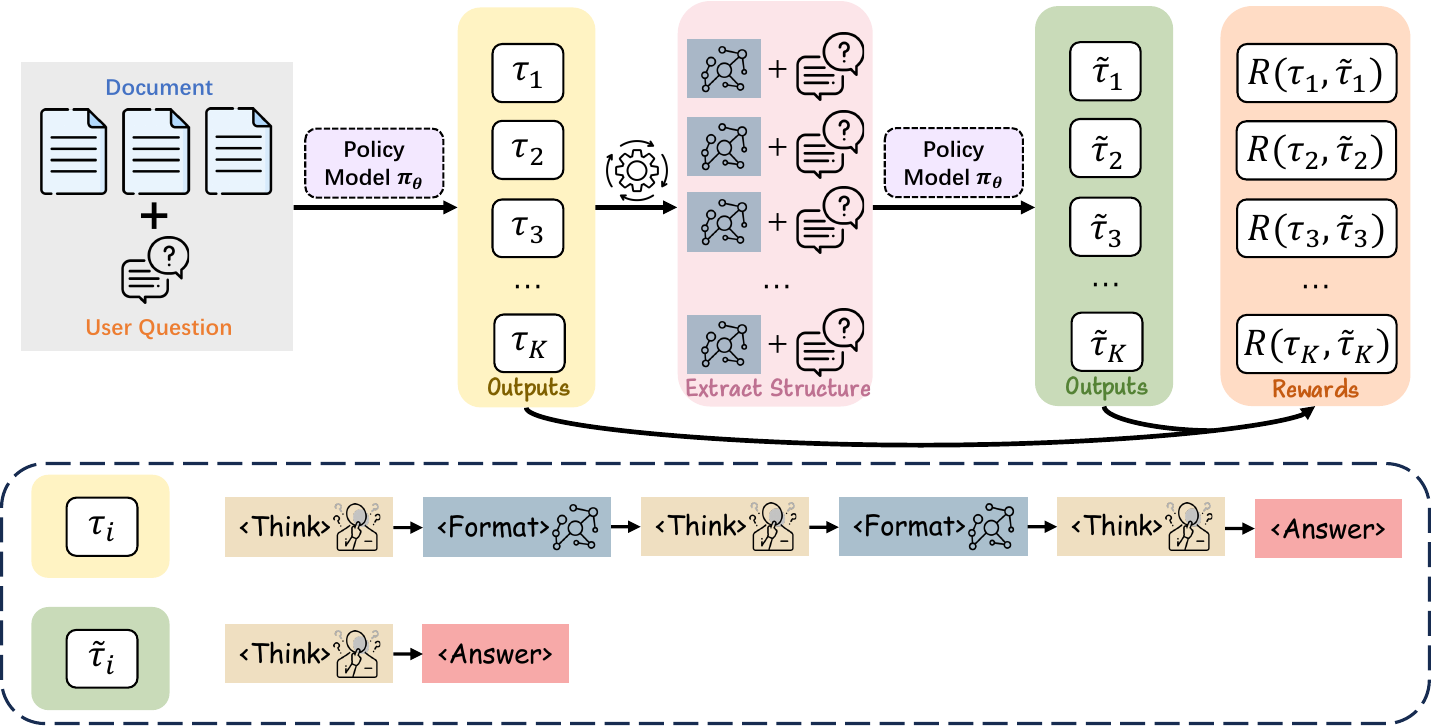}
    \caption{Reward calculation for GRPO training. The process evaluates the correctness of the generated answers and ensures that the extracted structures are self-contained, coherent, and informative.}
\label{fig:algorithm}
\end{figure*}

To promote explicit reasoning and structured formats transformation, 
we design a specialized prompt template guided by a set of fundamental rules. 
This template enables the model to answer a query $q$ using the retrieved documents $\mathcal{D}$. 
As illustrated in Figure~\ref{fig:prompt-template}, the prompt first instructs the language model to perform a \emph{think-and-structure} step, 
where it decomposes the content and organizes relevant information into a structured format. 
This process is repeated iteratively until the model determines that sufficient context and reasoning have been accumulated to provide a confident answer. 

At a high level, to ensure consistency and clarity in the reasoning process, we define a set of formatting rules: 
(1) The structured output must be enclosed within explicit tags following the format:
\begin{quote}
    \texttt{<format: format\_name>} … \texttt{</format: format\_name>}.
\end{quote}
This convention clearly delineates the beginning and end of each structured reasoning segment. 
(2) To encourage genuine and more sophisticated transformation from unstructured text to structured representations, the model is prohibited from directly copying or pasting descriptive content from the source documents into the structured format block. Instead, it must reformulate and abstract the relevant information through its own structured reasoning. The detailed of this prompting strategy is presented in Section~\ref{sec:templete_reasoning}.

\subsubsection{Design of Predefined Structural Formats} 

To facilitate the transformation of important information related to a query $q$ from the retrieved documents $\mathcal{D}$, we define a set of commonly used structural formats $\mathcal{S}$. These formats serve as templates to guide the LLM in efficiently selecting or generating suitable representations. The predefined structural formats include:
\begin{itemize}
    \item \texttt{Chunk}: A chunk is a self-contained summary of one or multiple documents in natural language.
    \item \texttt{Knowledge Graph}: A knowledge graph is a structured way of representation facts in the form of entities (things) and relations (connections between things), often expressed as triple: (head, relation, tail).
    \item \texttt{Table}: A table is a structured way of organizing data into rows and columns. It's commonly used to present information clearly and compactly.
    \item \texttt{Catalogue}: A catalogue is a structured, systematically arranged list of items-each described by a consistent set of metadata-that lets readers discover, browse, and retrieve individual entries quickly.
    \item \texttt{Algorithm}: An algorithm is a step-by-step procedure for solving a problem or achieving a specific result.
\end{itemize}

While the predefined structural formats $\mathcal{S}$ offer strong representations for many query types, they may not fully capture the diverse requirements of real-world reasoning tasks. In practice, it is unrealistic to assume that a fixed set of structural format schemas can cover all possible information needs. To address this challenge and take advantage of the generative capabilities of modern LLMs, we adopt an open-world setting where the model is encouraged to create new structural formats $\mathcal{S^+}$ when necessary. Specifically, we prompt the model with: 

\begin{center}
    \textit{You can also develop your own format if needed.}
\end{center}
Surprisingly, we find that \textsc{Structure-R1} is able to enrich the original format schema by inventing useful new structures tailored to specific queries. These dynamically generated formats contribute to better reasoning and lead to improved performance, as demonstrated by our experimental results in Section~\ref{sec:new_format}.

\subsubsection{Format-Aware Prompt Design for Structure-Enhanced Reasoning}
\label{sec:templete_reasoning}
Building on the structural format design, we introduce a format-aware prompting framework that enables large language models to perform multi-turn, structure-enhanced reasoning through explicit control over internal computation, intermediate context restructuring, and final answer generation.
This design transforms reasoning into a structured decision process, allowing the model to adaptively manage how information is processed, represented, and used. This design achieves two goals. First, the \emph{reasoning stage} 
(\texttt{<think> ... </think>}) forces the model to externalize intermediate thought before producing 
answers, aligning with chain-of-thought style supervision. Second, the \emph{formatting stage} 
(\texttt{<format: format\_name> ... </format: format\_name>}) provides a mechanism for the model 
to restructure the knowledge into more suitable representations (e.g., graphs, tables, catalogues, 
or algorithms). Finally, the \emph{answer stage} (\texttt{<answer> ... </answer>}) enforces a 
strictly delimited output. Together, these constraints define an agent-like policy space
where the model decides whether to continue reasoning, to transform context, or to finalize an answer. By enabling multiple cycles of \texttt{<think>} and \texttt{<format>} before \texttt{<answer>}, 
the model can adaptively transform knowledge representations and reassess their adequacy, 
rather than being constrained to a one-shot answer. 

\subsection{Outcome-directed Reinforcement Learning}
\subsubsection{Reinforcement Learning with Verifiable Reward}
We denote the backbone model as the policy model $\pi$, which is optimized using Reinforcement Learning with Verifiable Reward (RLVR). 
Given a query $q$, the policy model $\pi$ generates a reasoning trajectory $\tau_q$. 
For notational completeness, we also introduce $\tilde{\tau}_q$, which will later denote the re-inferred trajectory used for structural verification (Section~\ref{subsec_self_rwd}). 
Together, we refer to the pair as $t_q = (\tau_q, \tilde{\tau}_q)$. 

Our training framework for the policy model $\pi_\theta$ is based on the Group Relative Policy Optimization (GRPO) algorithm, which guides policy updates by comparing relative response quality within each group of sampled trajectories. 
For each query $q$, we sample a set of $K$ candidate response pairs, forming the set 
$\mathcal{T}_q = \{\,t_q^{(i)} = (\tau_q^{(i)}, \tilde{\tau}_q^{(i)})\,\}_{i=1}^K$, 
where each $t_q^{(i)} \sim \pi_\theta(\cdot \,|\, q)$. 
Each response $t_q^{(i)} \in \mathcal{T}_q$ is scored by a reward function $R(q, t_q^{(i)})$ 
based on whether the answers produced by $\tau_q^{(i)}$ and $\tilde{\tau}_q^{(i)}$ exactly match the ground truth. 
Since all responses in $\mathcal{T}_q$ correspond to the same query $q$, we simplify the notation to $R(t_q^{(i)})$. 
The group-normalized advantage is then computed as:
\begin{align*}
\hat{A}(t_q^{(i)}) 
    = R(t_q^{(i)}) 
    - \frac{1}{K} \sum_{j=1}^{K} R(t_q^{(j)}),
\end{align*}
where the reward is centered around the mean performance within the same group of sampled responses, encouraging the model to identify and reinforce relatively better outputs. 
The overall training objective is defined as:
\begin{align*}
\begin{split}
J(\theta) = 
\mathbb{E}_{t_q^{(i)} \sim \pi_\theta} 
\Big[ \,
& \min \Big( 
    \tfrac{\pi_\theta(t_q^{(i)})}{\pi_{\text{ref}}(t_q^{(i)})} 
    \, \hat{A}(t_q^{(i)}), \;
    \text{clip}\Big(
        \tfrac{\pi_\theta(t_q^{(i)})}{\pi_{\text{ref}}(t_q^{(i)})}, 
        1 \pm \epsilon
    \Big)
    \hat{A}(t_q^{(i)}) 
    \Big) \\
& - \beta \, D_{\mathrm{KL}}\big(\pi_\theta \,\|\, \pi_{\text{ref}}\big)
\Big],
\end{split}
\end{align*}
where $\hat{A}(t_q^{(i)})$ is the group-normalized advantage, 
and $\pi_{\text{ref}}$ is a reference model used for KL regularization, typically initialized as the frozen base model.  
Building on the format-aware prompt design introduced in Section~\ref{subsec_train_temp}, this training framework encourages the policy to leverage structured reformulations that enhance downstream reasoning performance, ultimately leading to more correct and verifiable answers.

\subsubsection{Warm-up Stage}
\label{subsub_warm}
As discussed earlier, our approach leverages structured representations to extract query-relevant information, thereby enhancing the reasoning capabilities of LLMs. 
To encourage this structure-to-reasoning behavior, we first perform a warm-up stage in which vanilla GRPO is applied to the \textsc{Structure-R1} model using our constructed prompt. 
During this stage, the model is trained with direct supervision on annotated $(\text{query}, \text{context}, \text{answer})$ triples, allowing it to learn how structured inputs can effectively guide reasoning. 
As mentioned in the previous section, the entire generated reasoning trajectory for a query $q$ is denoted by $\tau_q$. 
We use a simple rule-based reward function that evaluates only the final output, with correctness determined by exact string match, offering a clear and reliable training signal:
\begin{align*}
    R_{\text{direct}}(q, \tau_q) = \text{EM}(\hat{a}, a),
\end{align*}
where $\hat{a}$ denotes the model-generated answer for query $q$, and $a$ is the corresponding ground-truth answer. 
After this warm-up stage, the \textsc{Structure-R1} model achieves strong performance, demonstrating the ability to produce contextually grounded answers that leverage structured knowledge for improved accuracy.

\subsubsection{Self-Reward for Structural Verification}
\label{subsec_self_rwd}

Although the training approach described in Section~\ref{subsub_warm} equips LLMs with strong capability to transform plain text into structured formats, 
a key challenge remains: for the structured data to be truly useful in answering the query $q$, the resulting representations must be both correct and sufficiently self-contained. 
This limitation represents a major bottleneck that directly impacts the accuracy of the final generated answer $\hat{a}$.

To address this issue, we propose a self-reward module designed to verify and encourage the correctness and self-containment of the generated structured representations. 
Specifically, if a trajectory includes a transformation step 
(\texttt{<format: format\_name> ... </format: format\_name>}), 
we extract \textit{all} formatted information and perform a second inference conditioned on this structured representation. 
During this re-inference, the model receives only the structured information (i.e., without the original document context) together with the query question, following an R1-style prompt. 
This produces a new trajectory $\tilde{\tau}_q$ and an associated reward $R_{\text{reinf}}$. 
If no \texttt{<format>} block appears in the trajectory, we set $R_{\text{reinf}} = 0$, ensuring that re-inference contributes only when the model explicitly performs structured reformulation.

The prompt template used for this re-inference process is shown in Figure~\ref{fig:prompt-r1}. 
In this template, the \texttt{context} section is populated by concatenating the structured content extracted from all 
\texttt{<format: format\_name> … </format: format\_name>} blocks, separated by double newlines (\texttt{\textbackslash n\textbackslash n}). 
This re-inference step treats the structured information as a standalone context, enabling the model to generate an answer based solely on the reformatted knowledge.

\subsubsection{Reward Function Design.} 
The overall reward combines two components: 
(1) direct supervision ($R_{\text{direct}}$) and 
(2) self-reward for structural verification ($R_{\text{reinf}}$), 
and is defined as:
\begin{align*}
R(t_q) = R_{\text{direct}} + \lambda R_{\text{reinf}},
\end{align*}
where $\lambda$ is a weighting coefficient that balances the contributions of the direct inference reward and the re-inference reward. 
This design ensures that the policy trajectory receives additional reward when the structured content is both accurate and self-contained. 
In practice, $\lambda$ functions similarly to a learning rate and can be dynamically scheduled during training to modulate the influence of the re-inference signal over time.

\subsection{Theoretical Analysis}
In this section, we discuss the reasons why allowing LLMs to self-define the structures of external knowledge is the key to promoting their reasoning capability.

\subsubsection{Quantitative Metrics}
To establish a formal mathematical analysis framework, we first define scalar metrics to evaluate what makes \textsc{Structure-R1} better than the baselines.

Let $\mathcal{I}(a)$ denote the semantic information content contained in a token sequence $a$, measured in terms of relevant facts, entities, relationships, and logical structures necessary to answer a specific query $q$~\citep{shannon1948mathematical}. The \emph{information density} of $a$ is defined as: 
\begin{equation}
    \rho(a) = {\mathcal{I}(a)}/{|a|},
\end{equation}
which explicitly considers the number of tokens within $a$.

The structure transformation is defined as a mapping from the question $q\in\mathbb{Q}$ and external knowledge $\mathcal{D}_q\in 2^{\mathbb{D}}$ to a text-based format $s\in\mathbb{S}$: 
\begin{equation}
    \phi: \mathbb{Q} \times 2^{\mathbb{D}} \rightarrow \mathbb{S},
\end{equation}
and the optimal structure transformation of question $q$ should maximize the information density of the transformed structure: 
\begin{equation}
    \phi_q^\star = \argmax_{\phi}\rho(\phi(q, \mathcal{D}_q)).
\end{equation}

\subsubsection{\textsc{Structure-R1} Generates More Suitable Structures}
Built upon the information density, our analysis then reveals why \textsc{Structure-R1} can induce better structures with higher information density. We begin by introducing the optimal structures, which are considered the most suitable for the particular questions.

\begin{assumption}[Optimal Structure\label{assumption:fit_better}]
    For any question $q$, there exists one optimal structure $s_q^\star$ for representing the external knowledge $\mathcal{D}_q$, such that the information density is maximized: $s_q^\star = \argmax_{s}\rho(s)$.
\end{assumption}

Finding the optimal structure transformation for each question is a hard problem and requires extensive exploration via reinforcement learning~\citep{kaelbling1996reinforcement}, as demonstrated in Section~\ref{sec:method}. However, the potential of our proposed self-defined structures is provably better than a fixed structure set, as shown in Theorem~\ref{lemma:fit_better}.

\begin{theorem}[Self-Defined Structures Fit Better\label{lemma:fit_better}]
    Consider a predefined structure set $\mathcal{S}$. The information density upper bound associated with $\mathcal{S}$ is strictly no greater than that of self-defined structures:
    $$
    \rho(\mathcal{D}_q) < \max_{s\in\mathcal{S}}\rho(s) \leq \max_{\phi}\rho(\phi(q, \mathcal{D}_q)).
    $$
    The proof is detailed in Appendix~\ref{proof:fit_better}.
\end{theorem}

The right structure of a question preserves the semantic information and significantly reduces the token length, representing external knowledge in a minimal-noise manner.

\subsubsection{Better Structures Improve Reasoning Capability}
Another critical problem to answer is the relationship between suitable structures and the reasoning capability, as an explanation for the superior performance of \textsc{Structure-R1} on diverse benchmarks (Section~\ref{sec:exp}). Based on our empirical observation, we assume the following monotonic relationship between the expected information density and the accuracy.

\begin{assumption}[Performance Monotonicity\label{assumption:min_err}]
    Transformed structure $s$ with higher information density $\rho(s)$ leads to better reasoning capability (i.e., higher accuracy in the final answer). The classification error $\mathcal{E}(\pi_{\bm{\theta},\phi})$ of a LLM-based policy $\pi_{\bm{\theta},\phi}$ is monotonic to the information density:
    \begin{align*}
        &\mathcal{E}(\pi_{\bm{\theta},\phi_1}) < \mathcal{E}(\pi_{\bm{\theta},\phi_2}),~~~~\text{iff}\\
        &\mathbb{E}_{q,\mathcal{D}_q}\rho(\phi_1(q,\mathcal{D}_q)) > \mathbb{E}_{q,\mathcal{D}_q}\rho(\phi_2(q,\mathcal{D}_q)).
    \end{align*}
\end{assumption}

Then, we show that minimizing the classification error is equivalent to finding the optimal structure transformation $\phi_q^\star$ when the generation pattern of LLMs remains unchanged (Theorem~\ref{lemma:min_err}). This process can be automated via reinforcement learning, in which the final accuracy is used as the signal to find a better structure transformation $\phi$.

\begin{theorem}[Optimal Structure Means Minimal Error\label{lemma:min_err}]
    Under Assumption~\ref{assumption:min_err}, an optimal structure transformation $\phi_q^\star$ always leads to a minimal classification error:
    $$
    \argmin_{\phi}\mathcal{E}(\pi_{\bm{\theta},\phi}) \equiv \argmax_{\phi}\mathbb{E}_{q,\mathcal{D}_q}\rho(\phi(q, \mathcal{D}_q)).
    $$
    The proof is detailed in Appendix~\ref{proof:min_err}
\end{theorem}

The distinction of \textsc{Structure-R1} lies in its ability to learn high-information-density structures directly from the final accuracy. This end-to-end approach eliminates the need for structure design and improves reasoning capability simultaneously.

\section{Experiments\label{sec:exp}}
\begin{table*}[!htbp]
\footnotesize
\centering
\begin{tabular}{lcccc|*{5}{cc}}
\toprule
\textbf{Method} &
\multicolumn{2}{c}{HotpotQA} &
\multicolumn{2}{c|}{2Wiki.} &
\multicolumn{2}{c}{PopQA} &
\multicolumn{2}{c}{NQ} &
\multicolumn{2}{c}{TriviaQA} &
\multicolumn{2}{c}{Bamboogle} &
\multicolumn{2}{c}{Musique}\\
\cmidrule(lr){2-3} \cmidrule(lr){4-5} \cmidrule(lr){6-7} \cmidrule(lr){8-9}
\cmidrule(lr){10-11} \cmidrule(lr){12-13} \cmidrule(lr){14-15}  \
& EM & F1 & EM & F1 & EM & F1 & EM & F1 & EM & F1 & EM & F1 & EM & F1  \\
\midrule
\multicolumn{15}{c}{\textbf{GPT-4o-mini}}\\
GraphRAG 
& 36.2 & 49.42 
& 37.7 & 46.19 
& 38.09 & 46.66 
& 27.09 & 40.19 
& 62.70 & 72.24 
& 34.4 & 44.21 
& 10.65 & 20.44 \\
HippoRAG2 
& {37.90} & 52.8 
& 60.90 & 52.80 
& 41.2 & 50.04 
& 34.49 & 48.63 
& 60.30 & 71.41 
& 31.2 & 40.21 
& 14.77 & 24.77\\
HyperGraphRAG & 36.2 & 49.57 & 46.8 & 54.88 & 38.00& 47.45 & 26.70 & 38.46 & 60.80 & 71.14 & 23.2 & 32.34 & 10.40 & 19.80\\
StructRAG & 24.58 & 39.60 & 25.72 & 39.60 & 34.91 & 45.03 & 28.71 & 41.70 & 61.88 & 71.09 & 34.92 & 44.91 &  11.05 & 20.45\\
\midrule
\multicolumn{15}{c}{\textbf{Qwen2.5-72B-Instruct}}\\
GraphRAG  & 31.20 & 42.66 & 30.90 & 38.47 & 36.70 & 45.78 & 26.58 & 38.49 & 61.74 & 70.85 & 6.40 & 15.82 & 9.20 & 17.87 \\
HippoRAG2 & 37.20 & 48.47 & 55.50 & 63.63 & 41.10 & 49.74 & 36.50 & 48.15 & 61.90 & 71.91 & 29.60 & 35.86 & 11.70 & 19.81 \\
HyperGraphRAG  & 29.60 & 42.82 & 42.10 & 51.54 & 36.01 & 45.87 & 25.70 & 37.31 & 57.40 & 68.06 & 19.20& 27.03 & 9.00 & 17.40 \\
StructRAG & 27.45 & 41.84 & 24.54 & 37.18 & 36.46 & 45.03 & 27.88 & 40.56 & 60.80 & 71.93 & 44.82 & 54.48 & 8.78 & 17.17 \\
\midrule

\multicolumn{15}{c}{\textbf{Qwen2.5-7B-Instruct}}\\
Direct 
& 18.97 & 27.09 
& 24.98 & 30.07 
& 15.88 & 19.80 
& 14.98 & 24.35 
& 43.22 & 50.15 
& 8.80 & 17.17 
& 3.81 & 11.85\\
CoT 
& 8.40 & 12.99
& 11.79 & 15.09
& 11.87 & 16.44
& 10.14 & 19.26
& 41.76 & 49.69
& 15.20 & 20.77
& 1.70 & 5.13
\\
SFT 
& 20.54 & 29.63
& 28.12 & 32.45
& 14.76 & 18.67
& 16.70 & 25.88
& 42.65 & 49.93
& 12.80 & 18.95
& 4.80 & 12.97\\
RAG 
& 31.33 & 41.92
& 38.22 & 47.74
& 37.19 & 44.31
& 29.00 & 39.96
& 57.87 & 67.15
& 12.80 & 22.42
& 4.80 & 12.97\\
LongContext 
& 28.67 & 40.25
& 42.53 & 51.83
& 32.99 & 41.32
& 22.41 & 35.05
& 53.48 & 64.20
& 13.60 & 19.97
& 5.08 & 12.07
\\
GraphRAG 
& 22.00 & 33.59
& 30.48 & 37.83
& 12.80 & 27.39
& 7.30 & 21.41
& 34.50 & 47.99
& 8.00 & 14.93
& 4.10 & 11.83\\
HippoRAG & 18.20 & 32.15 & 19.70 & 38.22 & 26.80 & 37.00 & 10.30 & 25.11 & 37.00 & 51.04 & 15.20 & 23.66  & 4.43 & 12.84\\
HypergraphRAG 
& 4.40 & 6.24 
& 1.60 & 1.87
& 0.80 & 1.59
& 1.30 & 2.71
& 3.80 & 6.16 
& 3.20 & 3.89
& 0.36 & 1.18\\
StructRAG & 18.54 & 29.84 & 16.74 & 28.36 & 31.43 & 40.12 & 20.8 & 32.15 & 50.00 & 60.42 & 13.60 & 24.47 & 4.14 & 10.66\\
R1 & 37.46 & 48.39 & 69.78 & 77.20 & 40.97 & 47.58 & 34.60 & 46.20 & 60.50 & 69.05 & 24.00 & 31.75 & 7.61 & 15.79
\\
\textbf{Structure-R1 (Ours)} & \textbf{\underline{38.26}} & \textbf{\underline{49.67}} & \textbf{\underline{74.24}} & 
\textbf{\underline{79.95}} & 
\textbf{\underline{42.48}} & 
\textbf{48.48} & 
\textbf{\underline{37.67}} & \textbf{48.20} & \textbf{61.43} & 
\textbf{69.20} & 
\textbf{28.80} & 
\textbf{37.91} & 
\textbf{8.73} & 
\textbf{16.52}
\\
\bottomrule
\end{tabular}
\caption{Main Results. The bolded results represent the strongest performance among approaches that use Qwen2.5-7B-Instruct as the base model, while the underlined results indicate the strongest performance among all models.}
\vspace{-0.2in}
\label{tab:main_results}
\end{table*}

We evaluate our proposed method, \textsc{Structure-R1}, on seven knowledge-intensive reasoning benchmarks to address the following research questions (RQ):

\begin{itemize}
    \item  \textbf{RQ1:} How does \textsc{Structure-R1} perform compared to state-of-the-art baselines in knowledge-intensive reasoning tasks?
    
    \item \textbf{RQ2:} Which components of \textsc{Structure-R1} contribute most significantly to its effectiveness, and how can their associated hyperparameters be optimized?
    
    \item \textbf{RQ3:} How does dynamically expanding the structure format in the reasoning trajectory or growing the predefined schema during inference, impact performance on knowledge-intensive reasoning tasks?
\end{itemize}

\subsection{Experimental Settings}

\subsubsection{Datasets} 

We evaluate the performance of \textsc{Structure-R1} on seven benchmarks. As previously discussed, transforming unstructured retrieved documents into appropriate structured formats, and learning a dynamic policy to select or generate new structures at each reasoning step, can significantly enhance the reasoning capabilities of LLMs in knowledge-intensive, complex tasks. To assess this in the multi-hop reasoning setting, we select two in-domain benchmarks: HotpotQA~\cite{yang2018hotpotqa} and 2WikiMultiHopQA (2Wiki.)~\cite{ho2020constructing}. We sample 7,000 examples from the training sets of each benchmark for model training and evaluate on these two benchmarks for the performance on in-domain benchmarks. To assess generalization and domain adaptation capabilities, we further evaluate our method and baseline models on two out-of-domain multi-hop QA benchmarks: Bamboogle~\cite{press2023measuring} and MuSiQue~\cite{trivedi2022musique} and three Open-Domain Question Answering benchmarks: PopQA~\cite{mallen2023not}, Natural Questions (NQ)~\cite{kwiatkowski2019natural} and TriviaQA~\cite{joshi2017triviaqa}. Due to the large size of the test sets for HotpotQA, 2WikiMultiHopQA, TriviaQA, and PopQA, and given the limited budget when using proprietary models, we randomly sample 1,000 questions from each for evaluation. Detailed data statistics are provided in Table~\ref{tab:data_stats} (Appendix).

\subsubsection{Compared Methods.} In this work, our primary focus is on how to represent retrieved documents using suitable structured formats to help LLMs more effectively extract and utilize important information. Rather than improving retrieval performance through advanced indexing techniques or structural retrieval methods, our goal is to enhance the consumption of retrieved content by LLMs. Based on this objective, we compare three types of baseline methods. The first category includes methods that perform reasoning without access to retrieved documents:

\begin{itemize}
    \item \textbf{Direct Inference:} It prompts LLMs with the query question directly to generate answers.
    \item \textbf{Chain-of-Thought (CoT)~\cite{wei2022chain}:} It prompts LLM to  conduct reasoning step by step then answer the questions.
    \item \textbf{Supervised Fine-tuning (SFT) ~\cite{chung2024scaling}:} It adapts LLM to a specific task with labeled data.

\end{itemize}

The second category includes methods that perform reasoning with access to retrieved documents. It is important to note that our work does not aim to enhance retrieval effectiveness using structural information. To isolate the effect of different document representations, we fix the set of retrieved documents across all methods and focus solely on how the transformation of plain text into structured formats influences the reasoning performance.

\begin{itemize}
    \item \textbf{Retrieval Augmented Generation (RAG)~\cite{lewis2020retrieval}:} retrieves relevant text chunks based on text similarities between queries and documents, and appends relevant text chunks to given queries when answering questions.
    \item \textbf{GraphRAG~\cite{edge2024local}:} constructs knowledge graphs from documents and build community summaries to capture global relationships.
    \item \textbf{HippoRAG2~\cite{gutierrez2025rag}:} augments LLMs with a memory-like knowledge graph, combining Personalized PageRank and deeper passage integration.
    \item \textbf{HyperGraphRAG~\cite{luo2025hypergraphrag}:} extends graph-based RAGs to n-ary relational facts via hypergraphs.
    \item \textbf{StructRAG~\cite{listructrag}:} prompts LLMs to select a structure type from a given set for all documents relevant to a question, then prompts LLMs to transform documents into the chosen structure type, which are then appended to the questions.
\end{itemize}

The final category includes baseline methods that leverage verifiable rewards to fine-tune the LLM, enabling it to learn a reasoning policy for how to process and utilize the retrieved documents.
\begin{itemize}
    \item \textbf{R1~\cite{guo2025deepseek}:} It employs GRPO to learn the policy over plain-text retrieved documents. This approach does not involve any transformation of documents into structured formats; the LLM operates directly on unstructured text.
\end{itemize}

\subsubsection{Evaluation Metrics}

We evaluate the performance of proposed method and the baseline models on knowledge-intensive reasoning tasks using widely adopted metrics: Exact Match (EM) and F1-score. Exact Match (EM) measures the percentage of predictions that exactly match any of the gold answers. F1-score calculates the token-level harmonic mean of precision and recall between the predicted answer $\hat{a}_i$ and the reference answer $a_i$, capturing partial correctness. When multiple reference answers are provided, we compute the EM and F1 scores for each reference and report the maximum score for each instance. The formal definitions of these metrics are as follows:
\begin{align*}
    \text{EM} &= \frac{1}{N} \sum_{i=1}^{N} \mathbb{I} \left[ \text{norm}(\hat{a}_i) = \text{norm}(a_i) \right] \\
    \text{F1} &= \frac{1}{N} \sum_{i=1}^{N} \frac{2 \cdot | \text{tokens}(\hat{a}_i) \cap \text{tokens}(a_i)|}{|\text{tokens}(\hat{a}_i)| + |\text{tokens}(a_i)|}
\end{align*}
Here, $N$ is the number of examples. The function $\text{norm}(\cdot)$ applies answer normalization, and $\text{tokens}(\cdot)$ returns the set of tokens in the normalized string. $\mathbb{I}[\cdot]$ is the indicator function that returns 1 if prediction exactly matches the reference, and 0 otherwise.

\subsection{Result on Seven Knowledge-intensive Reasoning Benchmarks (RQ1)}

We compare \textsc{Structure-R1} against a range of state-of-the-art baselines across seven knowledge-intensive reasoning benchmarks: HotpotQA, 2Wiki, PopQA, NQ, TriviaQA, Bamboogle, and Musique, as shown in Table~\ref{tab:main_results}. The baselines include several structure-aware retrieval-augmented methods, such as GraphRAG, HippoRAG, HyperGraphRAG, and StructRAG. To assess whether our method effectively learns a policy from annotation data, we also include R1, a variant built on \textsc{Qwen2.5-7B-Instruct}, as an internal baseline. All models are evaluated under the same backbone architectures, including \textsc{Qwen2.5-7B-Instruct}, \textsc{Qwen2.5-72B-Instruct}, and \textsc{GPT-4o-mini}. Our key findings are summarized as follows:

First, \textsc{Structure-R1} consistently outperforms all baselines when using the same backbone model, \textsc{Qwen2.5-7B-Instruct}, across all benchmarks in both EM and F1 metrics. Notably, it surpasses the strong R1 baseline by up to 4.46 EM and 2.75 F1 on 2Wiki. On more complex, multi-hop datasets such as Bamboogle and NQ, \textsc{Structure-R1} also demonstrates substantial gains—outperforming R1 by 4.8 EM and 6.16 F1 on Bamboogle, and by 3.07 EM and 2.00 F1 on NQ. These improvements underscore the model’s ability to effectively integrate cross-hop evidence and structured knowledge, offering a clear advantage in scenarios where flat retrieval or naïve prompting strategies are insufficient for deep reasoning.

Second, despite being built on the smaller \textsc{Qwen2.5-7B-Instruct} backbone, \textsc{Structure-R1} often matches or even outperforms methods based on significantly larger models such as \textsc{GPT-4o-mini} and \textsc{Qwen2.5-72B-Instruct}. For example, it achieves the best performance across two in-domain benchmarks (HotpotQA and 2Wiki) and two out-of-domain benchmarks (PopQA and NQ). This demonstrates that structure-aware design can effectively compensate for limitations in model size. Specifically, on 2Wiki, \textsc{Structure-R1} improves Exact Match (EM) by 21.9\% and F1 by 25.6\% compared to the best-performing baselines using \textsc{GPT-4o-mini} and \textsc{Qwen2.5-7B-Instruct}, respectively. These results highlight that \textsc{Structure-R1} sets a new state of the art in knowledge-intensive reasoning, showcasing the effectiveness of structured modeling for aligning smaller models in complex question answering tasks.

\subsection{Investigating the Effect of Hyperparameter (RQ2)}

\begin{figure}[t]
\centering
\begin{subfigure}[t]{0.45\linewidth}
    \centering
    \includegraphics[width=\linewidth]{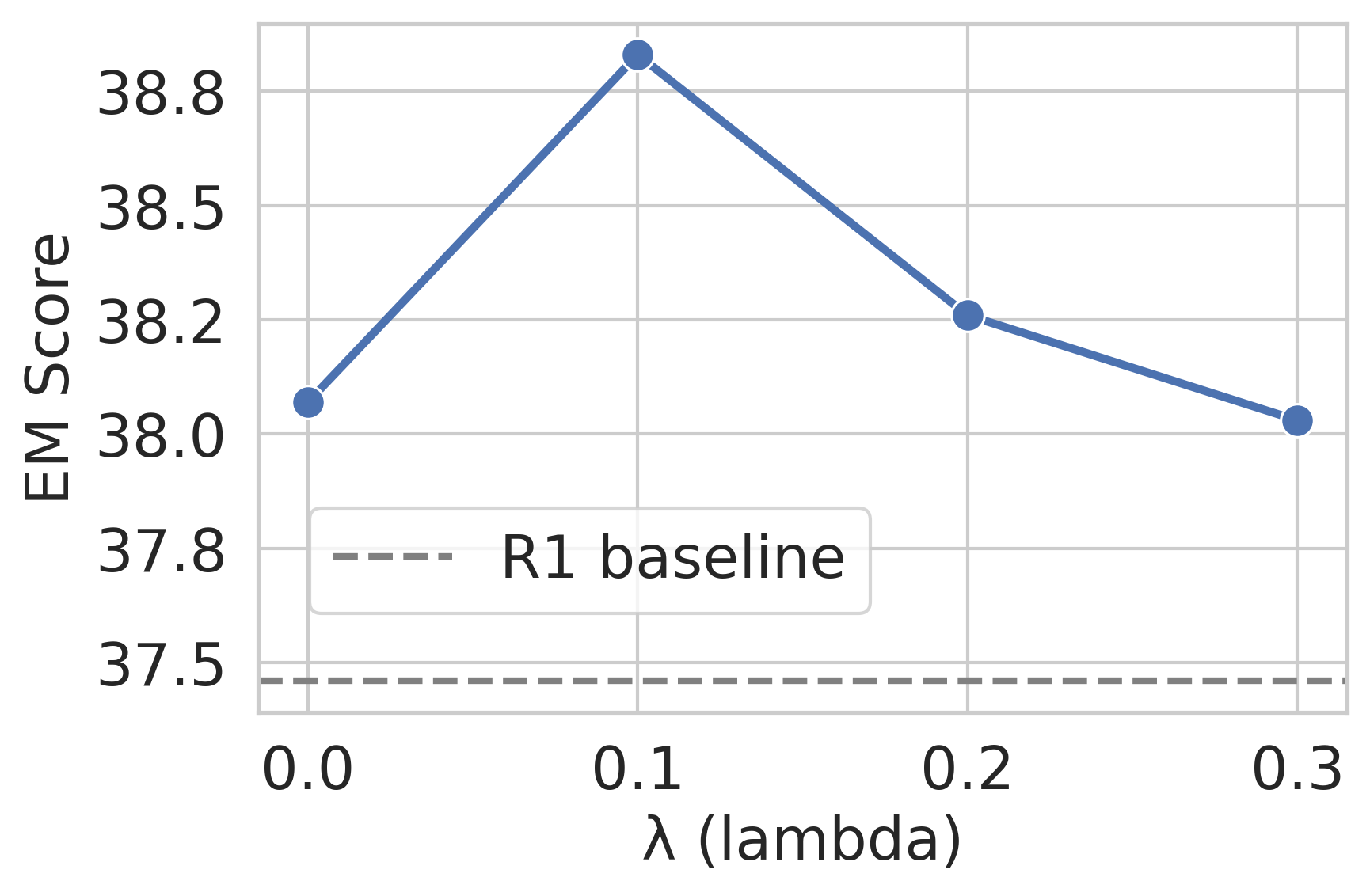}
\end{subfigure}
\hfill
\begin{subfigure}[t]{0.45\linewidth}
    \centering
    \includegraphics[width=\linewidth]{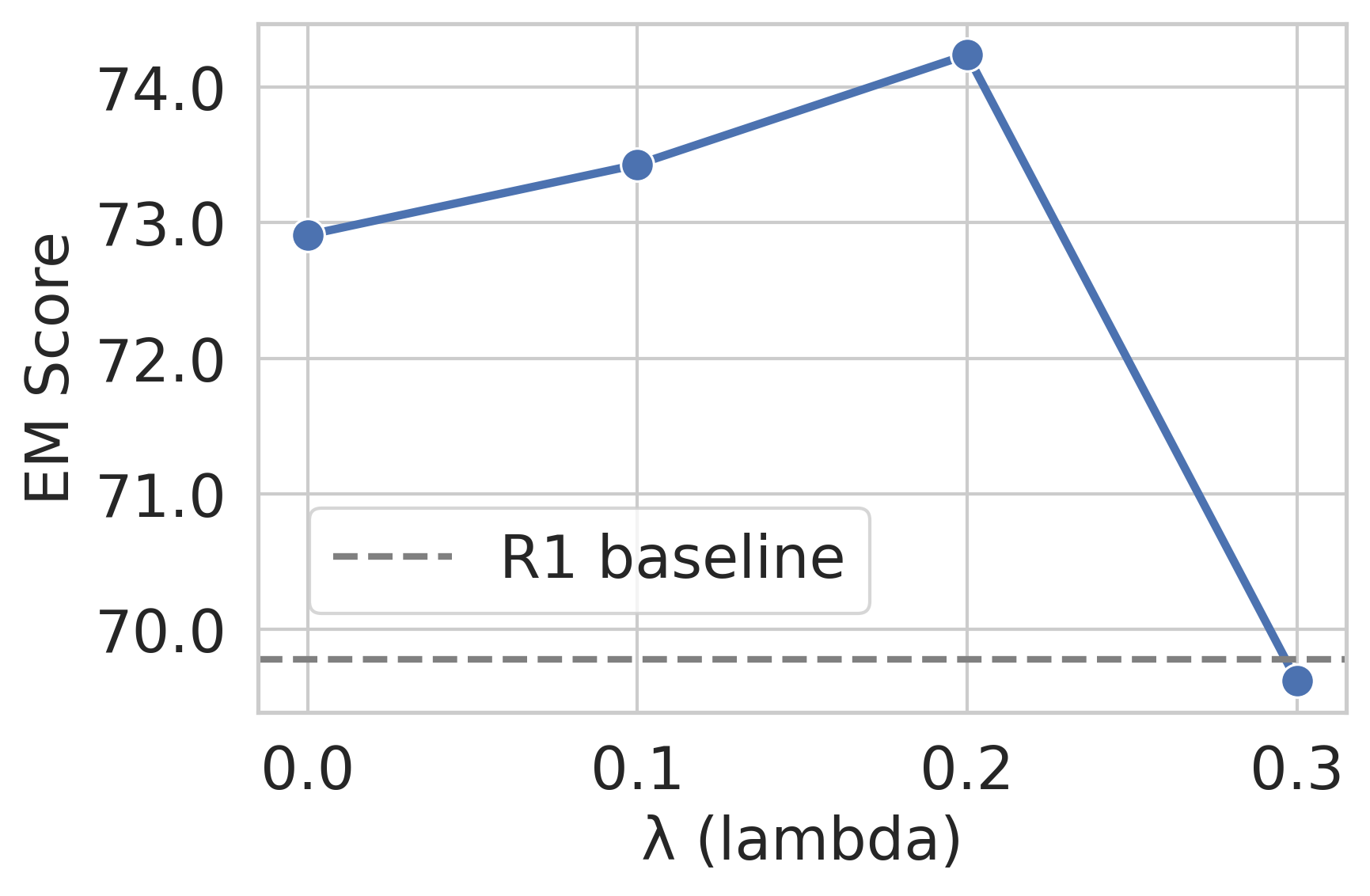}
\end{subfigure}


\begin{subfigure}[t]{0.45\linewidth}
    \centering
    \includegraphics[width=\linewidth]{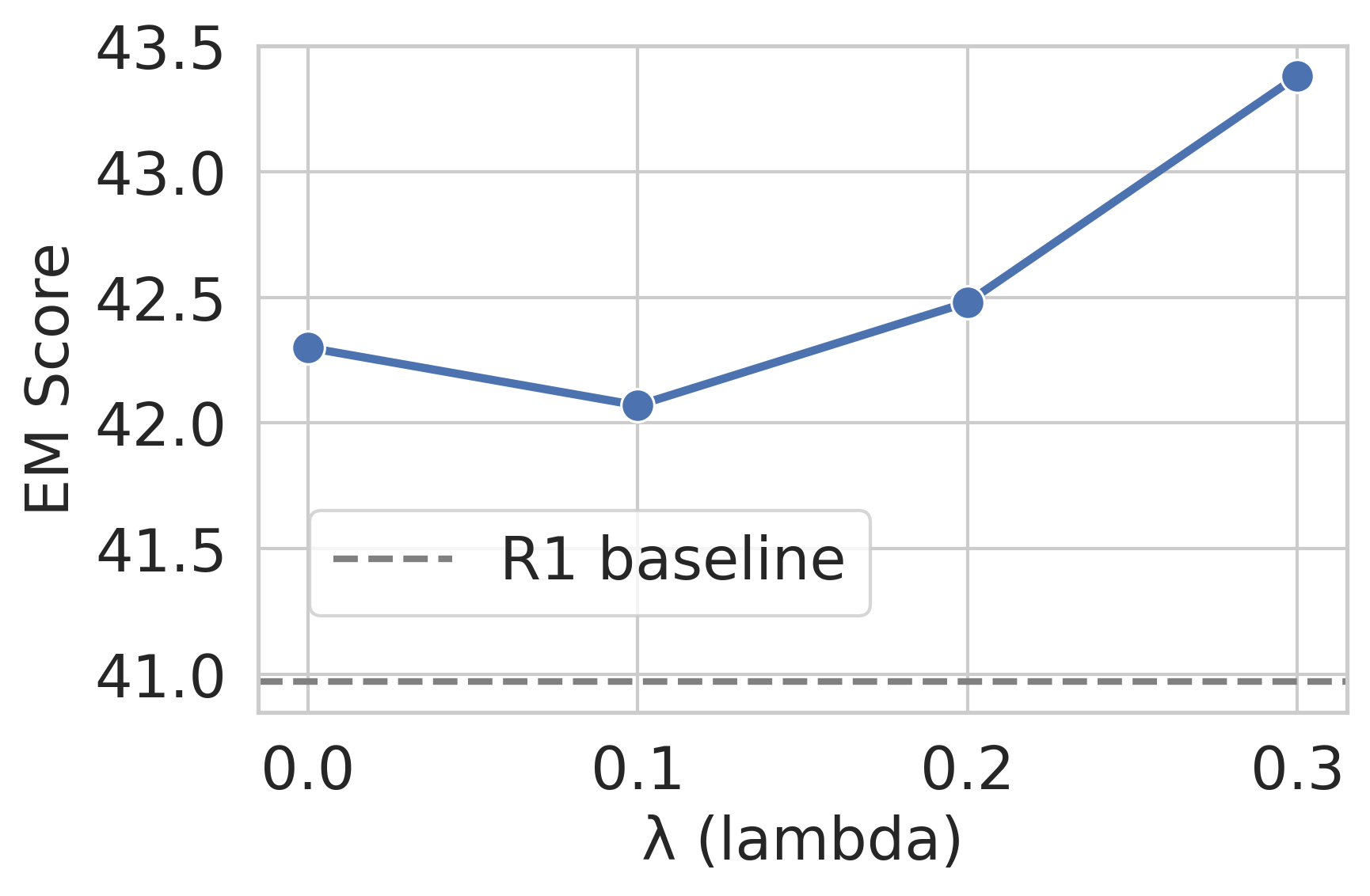}
\end{subfigure}
\hfill
\begin{subfigure}[t]{0.45\linewidth}
    \centering
    \includegraphics[width=\linewidth]{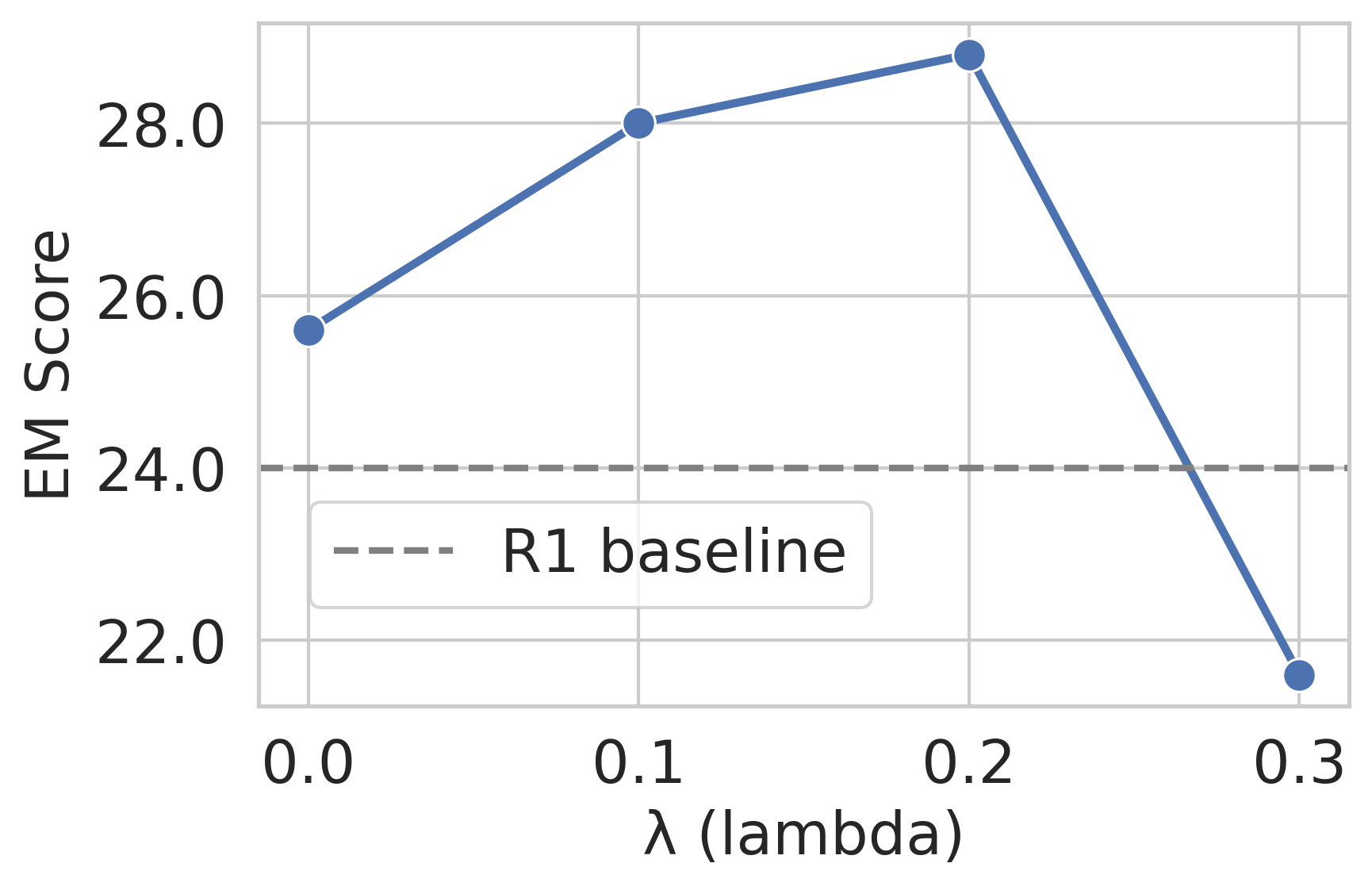}
\end{subfigure}
    \Description{}
\caption{Sensitivity analysis of model performance with respect to different values of $\lambda$. By averaging results across all datasets, we observe that the model achieves the most significant performance gain when $\lambda = 0.2$. Results are shown for HotpotQA (top left), 2Wiki. (top right), PopQA (bottom left), and Bamboogle (bottom right).}
\label{fig:dataset_grid}
\vspace{-0.2in}
\end{figure}

To better understand the impact of self-reward in the learning process, we conduct a sensitivity analysis on the reward weight $\lambda$, which controls the relative strength of the self-reward signal in our overall reward formulation. As $\lambda$ modulates the contribution of the re-inference reward $R_{\text{reinf}}$, tuning this parameter is crucial for balancing between direct supervision and structural self-correction. The results of this analysis are presented in Figure~\ref{fig:dataset_grid}, evaluated on two in-domain benchmarks (HotpotQA and 2Wiki.) and two out-of-domain benchmarks: Bamboogle for multi-hop QA and PopQA for open-domain QA.

From the results, we observe that a small $\lambda$ may underutilize the benefits of structure-guided feedback. Specifically, in both 2Wiki and Bamboogle, increasing $\lambda$ from 0.0 to 0.2 leads to a consistent improvement in the performance of our proposed method. However, setting $\lambda$ too high can overpower the supervised signal and destabilize the learning process. Except for PopQA, the performance on the other three benchmarks declines when $\lambda$ increases from 0.2 to 0.3. Notably, in Bamboogle and 2Wiki, our proposed method performs even worse than the general R1 baseline when $\lambda \geq 0.3$.  This suggests that while structured formats help aggregate key information, the semantic richness present in plain text remains essential for enhancing the model’s reasoning ability.

\subsection{Ablation Study (RQ2)}
We conduct an ablation study on four benchmarks (HotpotQA, 2Wiki, PopQA, and Bamboogle) to evaluate the contributions of two core components in the proposed \textsc{Structure-R1} framework: Multi-Structure Reasoning (Multi-Str.) and the Self-Rewarding mechanism (Self-Rwd.). As shown in Table~\ref{tab:domain-results}, enabling Multi-Str. consistently improves performance across all datasets, particularly on 2Wiki and Bamboogle, demonstrating its effectiveness in capturing multi-hop and structured dependencies. The most substantial improvements are observed when both components are enabled: for instance, on Bamboogle, the full configuration yields a +6.40 EM improvement over the baseline, with +3.20 contributed by Multi-Str. and +3.20 from Self-Rwd., highlighting their complementary strengths. The Self-Rwd. module further enhances robustness, especially in O.O.D. settings like PopQA and Bamboogle, where it provides a learning signal beyond supervised annotations. In contrast, its effect is less pronounced in in-domain datasets, suggesting that structured reasoning alone may be sufficient when task distribution aligns well with training data. These results verify that both components play vital roles in improving the effectiveness and generalization of \textsc{Structure-R1}.
ss diverse reasoning scenarios.

\begin{table}[t]
\centering
\footnotesize
\begin{tabular}{lcc|cccc}
\toprule
\textbf{Setting} & \multicolumn{2}{c|}{\textbf{Config}} &
\multicolumn{2}{c}{\textbf{HotpotQA}} &
\multicolumn{2}{c}{\textbf{2Wiki}} \\
\cmidrule(lr){2-3} \cmidrule(lr){4-5} \cmidrule(lr){6-7}
 & Multi-Str. & Self-Rwd. & EM & F1 & EM & F1 \\
\midrule
\multirow{3}{*}{I.D.}
 & \ding{55} & \ding{55} & 37.32 & 48.24 & 68.26 & 75.76 \\
 & \ding{51} & \ding{55} & 38.07 & 49.05 & 72.91 & 78.82 \\
 & \ding{51} & \ding{51} & \textbf{38.26 }& \textbf{49.67 }& \textbf{74.24} & \textbf{79.95} \\
 \bottomrule
\toprule
 \textbf{Setting}& \multicolumn{2}{c|}{\textbf{Config}} &
 \multicolumn{2}{c}{\textbf{PopQA}} &
 \multicolumn{2}{c}{\textbf{Bamboogle}} \\
 \cmidrule(lr){2-3} \cmidrule(lr){4-5} \cmidrule(lr){6-7}
 & Multi-Str. & Self-Rwd. & EM & F1 & EM & F1 \\
\midrule
\multirow{3}{*}{O.O.D.}
 & \ding{55} & \ding{55} & 42.35 & 48.22 & 22.40 & 31.56 \\
 & \ding{51} & \ding{55} & 42.30 & 48.45 & 25.60 & 35.44 \\
 & \ding{51} & \ding{51} & \textbf{42.48} & \textbf{48.48} & \textbf{28.80} & \textbf{37.91} \\
\bottomrule
\end{tabular}
\caption{Performance across in-domain (I.D.) and out-of-domain (O.O.D.) datasets under different configurations. ``Multi-Str.'' = Multi-Structure, ``Self-Rwd.'' = Self-Reward.\\ \ding{51}: feature enabled; \ding{55}: feature disabled.}
\label{tab:domain-results}
\vspace{-0.4in}
\end{table}

\subsection{Case Study (RQ3)}
\label{sec:new_format}

This section presents a case study demonstrating how \textsc{Structure-R1} can dynamically extend its predefined schema when the original structured format is not well-suited for the reasoning task. Given the question ``Which film has the director born later, The Girl in Possession or Así en el cielo como en la tierra?'', the model begins by extracting and organizing information using a table format. 

As shown in Table~\ref{tab:example_reasoning}, \textsc{Structure-R1} initially selects the table to list each film’s director along with their birth date. However, comparing birth dates is more nuanced than comparing numeric ages. It requires reasoning over the full date structure (year, month, and day). Recognizing this, \textsc{Structure-R1} adapts by introducing a more specialized format, \texttt{<date\_comparison>}, which reformats the dates into a normalized, machine-comparable form (e.g., converting month names into numerical values). This structural shift enables more accurate and transparent reasoning, ultimately guiding the model to the correct answer in a clearer and more effective way.

\begin{table}[t]
\footnotesize
\centering
\begin{tabular}{p{0.95\linewidth}}
\toprule
\textbf{Question:} Which film has the director born later, \emph{The Girl In Possession} or \emph{Así En El Cielo Como En La Tierra?} \\
\midrule
\begin{minipage}[t]{\linewidth}
\texttt{<think>}\\
To answer this question, I need to find the birth years of the directors of ``The Girl in Possession'' and ``Así en el cielo como en la tierra.''\\[0.3em]
From Doc~10, I can see that ``The Girl in Possession'' was directed by Monty Banks. From Doc~7, I know that Monty Banks was born Mario Bianchi on 15~July~1897.\\[0.3em]
From Doc~2, I can see that ``Así en el cielo como en la tierra'' was directed by José~Luis~Cuerda. From Doc~9, I know that José~Luis~Cuerda was born on 18~February~1947.\\[0.3em]
Now I need to compare the birth years of Monty~Banks and José~Luis~Cuerda.\\
\texttt{</think>}\\[0.3em]

\texttt{<format: table>}\\
My reformatted information:\\[0.1em]
\texttt{| Film Title | Director | Birth Date |}\\
\texttt{|------------|----------|------------|}\\
\texttt{| The Girl in Possession | Monty Banks | 15 July 1897 |}\\
\texttt{| Así en el cielo como en la tierra | José Luis Cuerda | 18 February 1947 |}\\
\texttt{</format: table>}\\[0.3em]

\texttt{<think>}\\
Now I have the birth dates of both directors in a table format. Monty~Banks was born on 15~July~1897, and José~Luis~Cuerda was born on 18~February~1947. \hlr{To determine which director was born later, I need to compare these dates.}\\
\texttt{</think>}\\[0.3em]

\texttt{\hls{<format: date\_comparison>}}\\
My reformatted information:\\
\texttt{- Monty Banks: 1897-07-15}\\
\texttt{- José Luis Cuerda: 1947-02-18}\\
\texttt{\hls{</format: date\_comparison>}}\\[0.3em]

\texttt{<think>}\\
Now that I have the dates in a comparable format, I can see that 1947 is later than 1897. Therefore, José~Luis~Cuerda was born later than Monty~Banks.\\
\texttt{</think>}\\[0.3em]

\hla{\texttt{<answer>} Así en el cielo como en la tierra \texttt{</answer>}}
\end{minipage}\\
\midrule
\textbf{Ground Truth:} \texttt{\{``target'': [``Así En El Cielo Como En La Tierra'', \hla{``Así en el cielo como en la tierra''}]\}} \\
\bottomrule
\end{tabular}
\caption{Illustration of a Structure-R1 reasoning trace. 
The model first organizes evidence into a table, then generates a novel 
\texttt{<format: date\_comparison>} structure to perform temporal reasoning beyond predefined schemas.}
\vspace{-0.4in}
\label{tab:example_reasoning}
\end{table}

\section{Related Work}
\noindent \textbf{Retrieval-Augmented Generation (RAG)}
Although large language models (LLMs) exhibit remarkable abilities in math reasoning and coding, they still suffer when facing domain-specific and knowledge-intensive tasks \cite{peng2023study, kandpal2023large}, and are prone to hallucinations \cite{huang2025survey}. Various retrieval-augmented generation (RAG) methods have been proposed to mitigate the aforementioned limitations. A RAG system typically consists of a large knowledge base, a retriever or search engine for retrieving relevant information, and a generator that integrates the question and retrieved context to produce an answer \cite{gao2023retrieval}. Standard RAG \cite{lewis2020retrieval, chung2024scaling} typically retrieves plain text passages based on text similarity and appends them to the input question for generation. However, plain text may contain irrelevant information that may mislead the generator. To better represent the knowledge, graph-based RAG \cite{edge2024local, gutierrez2025rag, listructrag} constructs a graph database by extracting knowledge graphs from texts, then a generator outputs an answer based on graph information, including graph community summary and retrieved knowledge graphs. However, graphs may not be sufficient to represent all kinds of data; for instance, a table may be more suitable for representing a company's revenue over the past five years. To improve upon graph-based retrieval-augmented generation (RAG), StructRAG \cite{listructrag} defines a set of structure types and selects one predefined structure format to represent contextual information. While we adopt the same benchmark settings as prior RAG work, our focus is fundamentally different—we aim to explore more effective ways of representing relevant texts for reasoning. To ensure a fair comparison and isolate the impact of representation, we fix the retrieved documents for all methods evaluated in this study.

\noindent \textbf{Reinforcement Learning with Verifiable Rewards (RLVR)}
Reinforcement learning (RL)~\citep{kaelbling1996reinforcement} has been extensively used in the post-training of LLMs, in which LLMs generate a set of rollout responses (exploration) and then enforce the rewarded responses while punishing the less-rewarded ones (exploitation). How to define the reward scores of LLMs' responses is a critical challenge. Reinforcement Learning from human feedback (RLHF)~\citep{ouyang2022training} explicitly learns reward models (RMs) as a proxy of human preference. However, RM-based reward has been shown to suffer from significant reward hacking problems~\citep{weng2024rewardhack}, which makes the LLMs poorly generalized to new questions. To circumvent this limitation, researchers turn to the reasoning tasks with hard-coded rules (e.g., math solving~\citep{cobbe2021gsm8k, grotschla2025benchmarking} and code generation~\citep{chen2021codex, jain2024livecodebench}). On these verifiable tasks, RL significantly improves the reasoning capability of LLMs~\citep{shao2024deepseekmath}. Recent breakthroughs of RLVR include the introduction of tool-calling~\citep{jin2025search, feng2025retool, qian2025toolrl}, multi-turn generation~\citep{shani2024multi, zhouarcher, zhou2025sweet}, and efficient rollout strategies~\citep{li2024cascade, yao2025flashrl, li2025reward}.

\section{Conclusion}
In this paper, we introduce \textsc{Structure-R1}, a novel framework for knowledge-intensive reasoning that learns a policy to generate structured representations from plain-text contextual information. Our method leverages reinforcement learning to enable the LLM to dynamically generate and expand structure formats, selecting the most appropriate representation based on the query. Extensive experiments on seven benchmarks demonstrate that \textsc{Structure-R1} achieves state-of-the-art performance among models using a 7B-scale backbone, and even outperforms larger models such as GPT-4o-mini on several tasks.


\bibliographystyle{ACM-Reference-Format}
\bibliography{main}

\newpage
\appendix

\section{Proofs}

\subsection{Proof of Theorem~\ref{lemma:fit_better}\label{proof:fit_better}}
\emph{Proof.} We first prove the LHS of Theorem~\ref{lemma:fit_better}. The original textual knowledge $\mathcal{D}_q$ is considered complete and extremely lengthy, while the predefined best structure $s_q^\star = \argmax_{s\in\mathcal{S}}\rho(s)$ is considered to preserve most of the information and is much shorter:
\begin{equation}
    \mathcal{I}(\mathcal{D}_q) \approx \mathcal{I}(s_q^\star)~~~~\text{and}~~~~|\mathcal{D}_q| \gg |s_q^\star|.
\end{equation}
Therefore, the inequality in the LHS holds:
\begin{equation}
    \rho(\mathcal{D}_q) = \frac{\mathcal{I}(\mathcal{D}_q)}{|\mathcal{D}_q|} < \frac{\mathcal{I}(s_q^\star)}{|s_q^\star|} = \rho(s_q^\star) = \max_{s\in\mathcal{S}}\rho(s).
\end{equation}

Then, we prove the RHS of Theorem~\ref{lemma:fit_better}. Given that the structure transformation $\phi$ is also implemented on the LLMs, it is normal to assume that all predefined structures can be learned by the transformation $\phi$:
\begin{equation}
    \forall s\in\mathcal{S}, \exists\phi, s = \phi(q, \mathcal{D}_q).
\end{equation}
When the best structure within $\mathcal{S}$ is different from the global optimal structure, the structure transformation $\phi$ can still generate a better one than all structures in $\mathcal{S}$. Taking the above two cases together, we have:
\begin{equation}
    \forall s\in\mathcal{S}, \exists\phi, \rho(\phi(q,\mathcal{D}_q)) \geq \rho(s).
\end{equation}
Therefore, we can always find a particular structure transformation $\phi$ to satisfy the RHS. \qed

\subsection{Proof of Theorem~\ref{lemma:min_err}\label{proof:min_err}}
\emph{Proof.} Consider the optimal structure transformation $\phi^\star$ w.r.t. the expected information density:
\begin{equation}
    \phi^\star = \argmax_{\phi}\mathbb{E}_{q,\mathcal{D}_q}\rho(\phi(q, \mathcal{D}_q)).
\end{equation}
Assume there exists a different $\phi'$ that achieve smaller classification error: $\mathcal{E}(\pi_{\bm{\theta},\phi'}) < \mathcal{E}(\pi_{\bm{\theta},\phi^\star})$. Then, the information density must satisfy:
\begin{equation}
    \mathbb{E}_{q,\mathcal{D}_q}\rho(\phi'(q,\mathcal{D}_q)) > \mathbb{E}_{q,\mathcal{D}_q}\rho(\phi^\star(q,\mathcal{D}_q)),
\end{equation}
which contradicts with the condition that $\phi^\star$ is optimal. \qed

\section{Prompt Templates}
Figure~\ref{fig:prompt-template} and~\ref{fig:prompt-r1} shows the prompt templates used in our \textsc{Structure-R1} framework.
\begin{figure*}[t]
\centering
\begin{minipage}{0.97\textwidth}
\begin{lstlisting}
Answer the given question. 
Related information is provided in <context> ... </context>. 
You must first conduct reasoning inside <think> ... </think>. 
After reasoning, if you find the provided information is hard to answer the question, you can transform the provided information into a more suitable format by <format: format_name> Your reformatted information </format: format_name>. 
Every time after you transform the provided information into a more suitable format, you must first conduct reasoning inside <think> ... </think>. 

!!! STRICT FORMAT RULES for <format: format_name>: !!!
    + You MUST replace format_name with the real format name, e.g. graph, table, algorithm, etc. 
    + You MUST replace Your reformatted information with a CONCRETE reformatted information that helps answer the original question below. 
    + NEVER copy or paste model descriptions into <format: format_name>. 
    + NEVER output the placeholder format <format: format_name> Your reformatted information </format: format_name>. Always replace both parts correctly. 

### The Descriptions of each Format:

Chunk: 
A chunk is a self-contained summary of one or multiple documents in natural language.

Knowledge Graph: 
A knowledge graph is a structured representation of facts in the form of entities (things) and relations (connections between things), often expressed as triples: (head, relation, tail).

Table: 
A table is a structured way of organizing data into rows and columns. It's commonly used to present information clearly and compactly.

Catalogue: 
A catalogue is a structured, systematically arranged list of items-each described by a consistent set of metadata-that lets readers discover, browse, and retrieve individual entries quickly.

Algorithm: 
An algorithm is a step-by-step procedure for solving a problem or achieving a specific result.

**You can also develop your own format by yourself.** Different documents might need different formats.

If you are ready to answer the question, you can directly provide your final answer inside <answer> ... </answer>, without additional explanation or illustration. 
For example: <answer> Beijing </answer>. 
    + Important: You must not output the placeholder text "<answer> and </answer>" alone. 
    + You must insert your actual answer between <answer> and </answer>, following the correct format. 

<context>
{context}
</context>

Question: {question}\n
\end{lstlisting}
\end{minipage}
\caption{Prompt template used in our \textsc{Structure-R1} framework.}
\label{fig:prompt-template}
\end{figure*}

\begin{figure*}[t]
\centering
\begin{minipage}{0.97\textwidth}
\begin{lstlisting}
Use the context below to answer the question as clearly and accurately as possible.

<context>
{context}
</context>

Question: {question}

Respond with one or more reasoning steps inside <think>...</think> tags.  
Then give the final answer inside an <answer>...</answer> tag.  
Avoid unnecessary repetition or placeholder content.

Example format:

<think>
Your reasoning goes here.
</think>

<answer>
Your final answer.
</answer>
\end{lstlisting}
\end{minipage}
\caption{R1-like prompt template used in our \textsc{Structure-R1} framework and the R1 baseline.}
\label{fig:prompt-r1}
\end{figure*}

\begin{table}[htbp]
\footnotesize
\begin{tabular}{c|ccc}
\toprule
  & Training & Testing & Sampled Testing \\
\midrule
HotpotQA   & 7,000    & 7,405  & 1,000 \\
2Wiki & 7,000    & 12,576 & 1,000 \\
\midrule
PopQA & -    & 14,267 &  1,000\\
NQ & -    & 3,610 & 1,000 \\
TriviaQA & -   & 11,313 & 1,000 \\
Bamboogle & -   & 125 & 125 \\
Musique & -    & 2,417 & 1,000  \\
\bottomrule
\end{tabular}
\caption{Statistics of the seven benchmarks used in this study. We trained our proposed methods and baselines on 7,000 examples sampled from HotpotQA and 2Wiki. For open-source backbone models at the 7B scale, we evaluated them on the full test sets . For proprietary models used in comparison, due to cost considerations, we randomly sampled 1,000 test examples when the full test set exceeded that size.}
\label{tab:data_stats}
\end{table}

\section{Statistics of benchmarks}
Table~\ref{tab:data_stats} shows the statistics of the seven benchmarks used in the paper.

\end{document}